\documentclass[letterpaper, 10 pt, conference]{ieeeconf}  

\IEEEoverridecommandlockouts                              

\usepackage{graphicx}
\usepackage{utfsym}
\usepackage{biblatex}  
\usepackage{multirow}
\usepackage{xcolor}

\title{\LARGE \bf
Dynamic Interactive Relation Capturing via Scene Graph Learning \\for Robotic Surgical Report Generation\
}

\author{Hongqiu Wang$^{1}$, Yueming Jin$^{2}$, and Lei Zhu$^{3}$ 
\thanks{The work was supported by the National Natural Science Foundation of China (Project No. 61902275), and  the Wellcome/EPSRC Centre for Interventional and Surgical Sciences (WEISS) [203145/Z/16/Z].}
\thanks{$^{1}$Hongqiu Wang is with The Hong Kong University of Science and Technology (Guangzhou), Nansha, Guangzhou, 511400, Guangdong, China.
        {\tt\small hwang007@connect.hkust-gz.edu.cn}}%
\thanks{$^{2}$Yueming Jin is with Wellcome/EPSRC Centre for Interventional and Surgical Sciences (WEISS) and Department of Computer Science, University College London
        {\tt\small yueming.jin@ucl.ac.uk}}%
\thanks{$^{3}$Lei Zhu is with  The Hong Kong University of Science and Technology (Guangzhou), Nansha, Guangzhou, 511400, Guangdong, China
and The Hong Kong University of Science and Technology, Hong Kong SAR, China.
        {\tt\small leizhu@ust.hk}}%
\thanks{$^{4}$Lei Zhu (leizhu@ust.hk) is the corresponding author of this work.}
}

\begin{document}

\maketitle
\thispagestyle{empty}
\pagestyle{empty}

\begin{abstract}

For robot-assisted surgery, an accurate surgical report reflects clinical operations during surgery and helps document entry tasks, post-operative analysis and follow-up treatment. 
It is a challenging task due to many complex and diverse interactions between instruments and tissues in the surgical scene. 
Although existing surgical report generation methods based on deep learning have achieved large success, they often ignore the interactive relation between tissues and instrumental tools, thereby degrading the report generation performance.
This paper presents a neural network to boost surgical report generation by explicitly exploring the interactive relation between tissues and surgical instruments.
To do so, we first devise a relational exploration (RE) module to model the interactive relation via graph learning, and an interaction perception (IP) module to assist the graph learning in RE module.
In our IP module, we first devise a node tracking system to identify and append missing graph nodes of the current video frame for constructing graphs at RE module. 
Moreover, the IP module generates a global attention model to indicate the existence of the interactive relation on the whole scene of the current video frame to eliminate the graph learning at the current video frame.
Furthermore, our IP module predicts a local attention model to more accurately identify the interaction relation of each graph node for assisting the graph updating at the RE module.  
After that, we concatenate features of all graph nodes of RE module and pass concatenated features into a transformer for generating the output surgical report.
We validate the effectiveness of our method on a widely-used robotic surgery benchmark dataset, and experimental results show that our network can significantly outperform existing state-of-the-art surgical report generation methods (e.g., 7.48\% and 5.43\% higher for BLEU-1 and ROUGE). 

\end{abstract}

\section{INTRODUCTION}

Robot-Assisted Minimally Invasive Surgery (RAMIS) has shown increasingly essential in recent decades given its several advantages, such as high stability, superhuman dexterity and intelligence \cite{2} \cite{add1}. RAMIS can bring great benefits to patients with reduced recovery time and trauma after surgery \cite{1}. Conventionally, surgeons need to generate a corresponding surgical report to record the surgical procedure performed by the surgical robots. It can provide a detailed reference for post-operative analysis of the surgical interventions \cite{9}. However, this task is generally time-consuming and labor-intensive. In this regard, automatic surgical report generation is highly demanded to reduce the burden of surgeons from low-level documentation task, allowing them to pay more attention to post-operative analysis on patients \cite{8}. 
Surgical report generation can also be seen as image caption generation\cite{3}, a composite task involving Computer Vision (CV) and Natural Language Processing (NLP) \cite{4}.

Image caption task transforms visual features extracted by the Convolutional Neural Networks (CNNs) into high-level semantic information. It is a complicated problem since it includes the detection of objects in images, understanding the inter-relationships between main objects, and finally expressing them in reasonable language. In the medical field, most research on diagnostic report generation has focused on medical images rather than surgical videos, such as radiology and pathology images \cite{7} \cite{jing}. However, with the development of RAMIS, the generation of surgical reports has received more and more attention, and there are a few latest papers in this field \cite{9} \cite{8}. Compared with diagnostic report generation, surgical report generation not only needs to describe the surgical instruments that appear in the surgical scene but also needs to pay attention to the interaction between instruments and tissues. Therefore, it requires a deeper understanding of the relationship between objects.

\begin{figure}[t]
    \centering
    \includegraphics[width=0.5\textwidth]{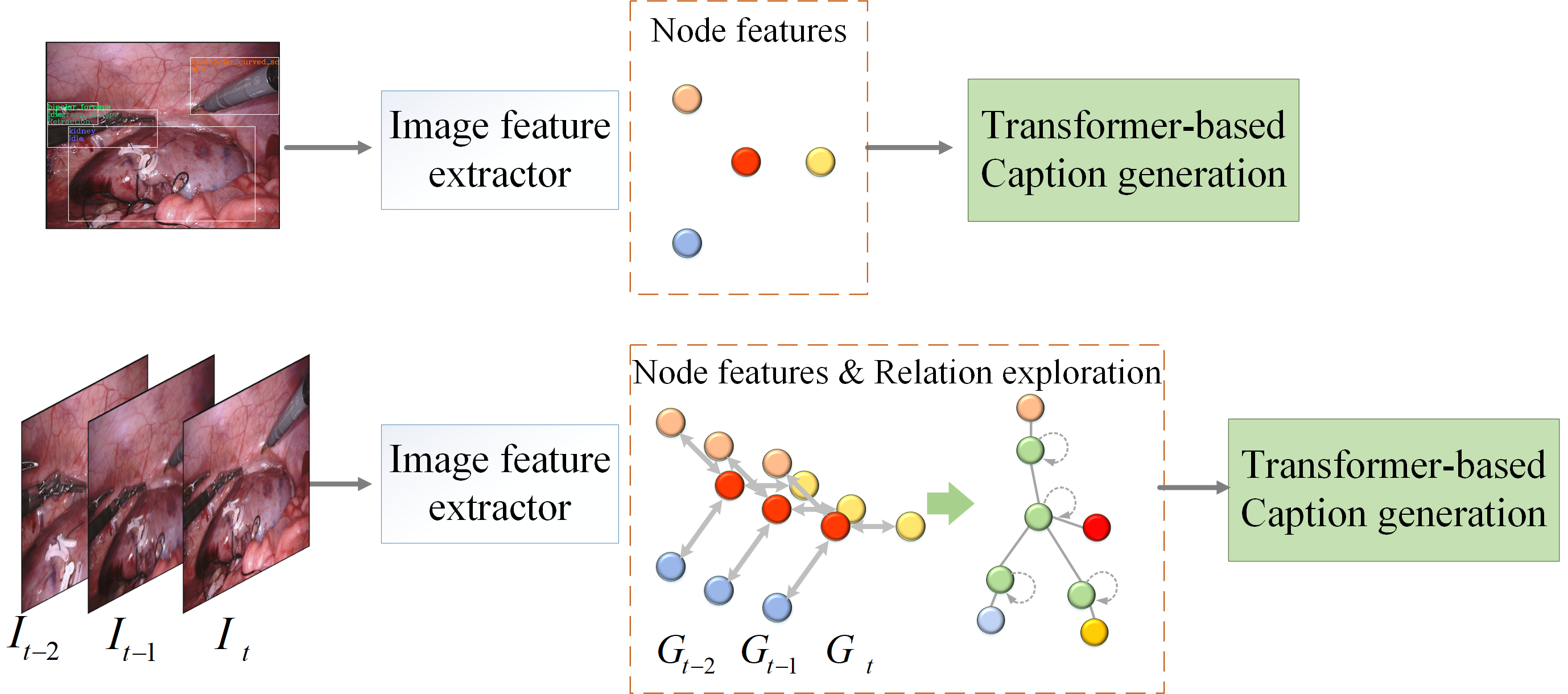}
    \caption{An overview of our proposed method (second row), against the conventional approach (first row) of deep learning pipelines for surgical report generation.}
    \label{fig:compare}
\end{figure}

\begin{figure*}[h]
    \centering
    \includegraphics[width=1\textwidth]{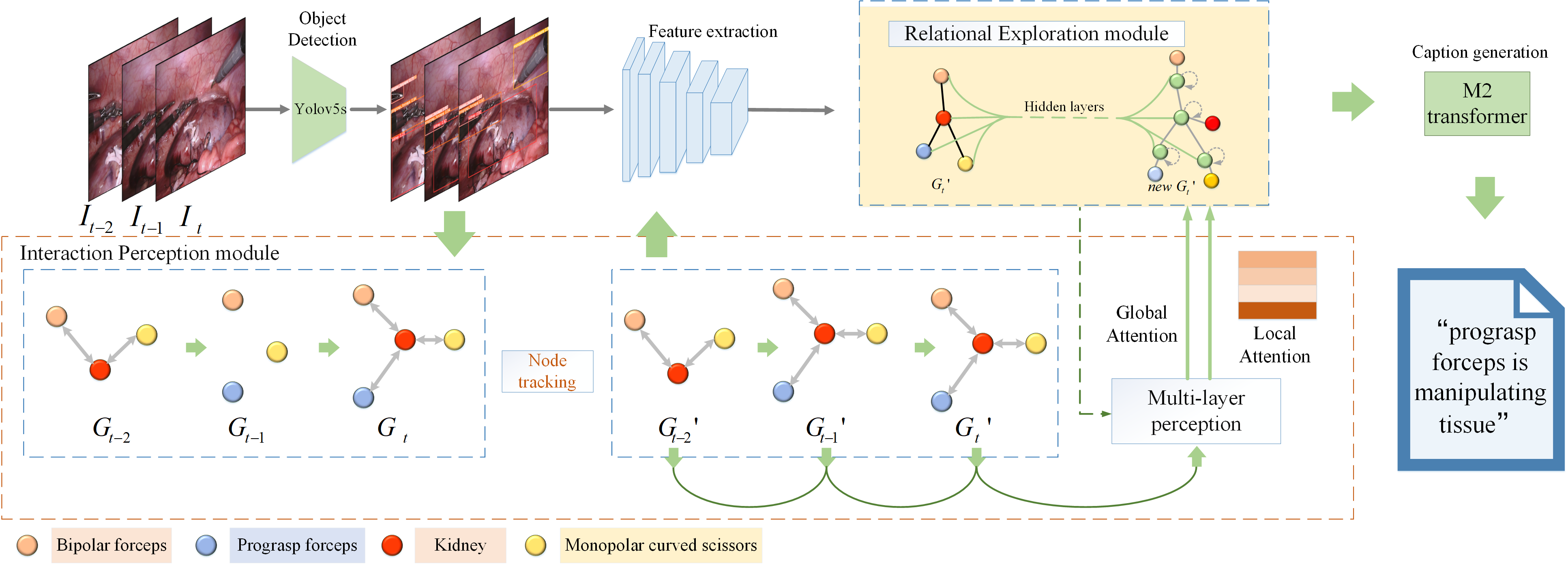}
    \caption{The architecture of the proposed model. The network includes a region feature extraction part, a relational exploration module, an interaction perception module and a caption generation part. It takes successive video frames as input and generates corresponding surgical reports.}
    \label{fig:pipeline}
\end{figure*}

Earlier methods for tackling image caption in the medical domain utilize CNN and long-short term memory (LSTM) network, to take advantage of high-level spatial temporal feature extraction \cite{10,11}. However, they suffer from limited representational abilities and generally encounter optimization difficulties. Recently, Transformer \cite{12} has made great successes in caption generation tasks of natural images \cite{13} \cite{14}, given its discriminative representation capability with self attention mechanism. Considering the excellent performance, it is also adopted as the main captioning architecture in surgical report generation \cite{9} \cite{8}. 
Most current works focus on the problem of domain adaptation \cite{9} \cite{8}, mainly for considering that there are new instruments and variations in surgical tissues appearing in robotic surgery. For example, Xu et al. \cite{8} propose the gradient reversal adversarial learning scheme, the gradient multiplies with a negative constant and updates adversarially in backward propagation, discriminating between the source and target domains and emerging domain-invariant features. Eventually, these image features are converted into text representations via the transformer. Additionally, a paper \cite{31} argues that mainstream captioning models still rely on object detectors or feature extractors to extract regional features. Therefore, they design an end-to-end detector and a feature extractor-free captioning model to simplify the process using the patch-based shifting window technique.

Although the current methods have achieved relatively good results, there are three points that can be improved. Firstly, various complex interactive relationships between instruments and tissues are important components for surgical report generation, while current methods have not explored the interactions between objects. Secondly, the current methods use a single frame of the surgical video as input to generate a report. However, considering that robotic surgery is a continuous process, temporal information is supposed to be reasonably utilized to facilitate task performance. Thirdly, most of them require additional bounding box information as input, while such annotations are expensive and inputting raw images is more practical.

Recently, graph neural networks (GNNs) have received increasing research interest because of their ability to learn non-Euclidean relations between entities \cite{gnn1}  \cite{15} \cite{16}. Many underlying complex relationships among data in several areas of science and engineering, e.g., computer vision, molecular biology, and pattern recognition, can be represented in terms of graphs \cite{gnn2}. GNN is widely used in the above fields and has achieved good performance \cite{gnn3} \cite{gnn4}. These achievements motivate us to utilize graph learning to explore the interaction between different nodes in the robotic surgery scene graph.

To alleviate the above issues, this paper proposes the relational exploration (RE) module that allows the network to perform spatial reasoning based on features extracted from the nodes of the scene graph (as shown in Fig. 1). Besides, interaction perception (IP) module is developed to apply temporal information and combine scene graph information to learn the interactive situation of the current video frame. It can generate global attention for the RE module to decide whether model the relation between different nodes and generate local attention maps to strengthen important nodes and suppress non-interactive nodes. Moreover, an object detector is applied to the raw image to replace the input and this seems also feasible from the experimental results.

Main contributions of this study are summarized as follows:
\begin{itemize}

\item We devise a graph learning framework for boosting surgical report generation via interactive relation reasoning along temporal dimension.

\item We propose a RE module that can learn interactive relationships between the tissue and instruments in the non-Euclidean domain to improve the accuracy of surgical report generation.
\item To serve this task well with temporal information, we devise an IP module to utilize both temporal information and scene graph information to focus on important interactions and nodes.
\item Experimental results on benchmark datasets show that our network clearly outperforms state-of-the-art surgical report generation methods. Even though our method does not take object bounding box as the input, our network still outperforms state-of-the-art methods, which utilizes the object bounding boxes as the input.
\end{itemize}



\begin{figure*}[h]
    \centering
    \includegraphics[width=0.86\textwidth]{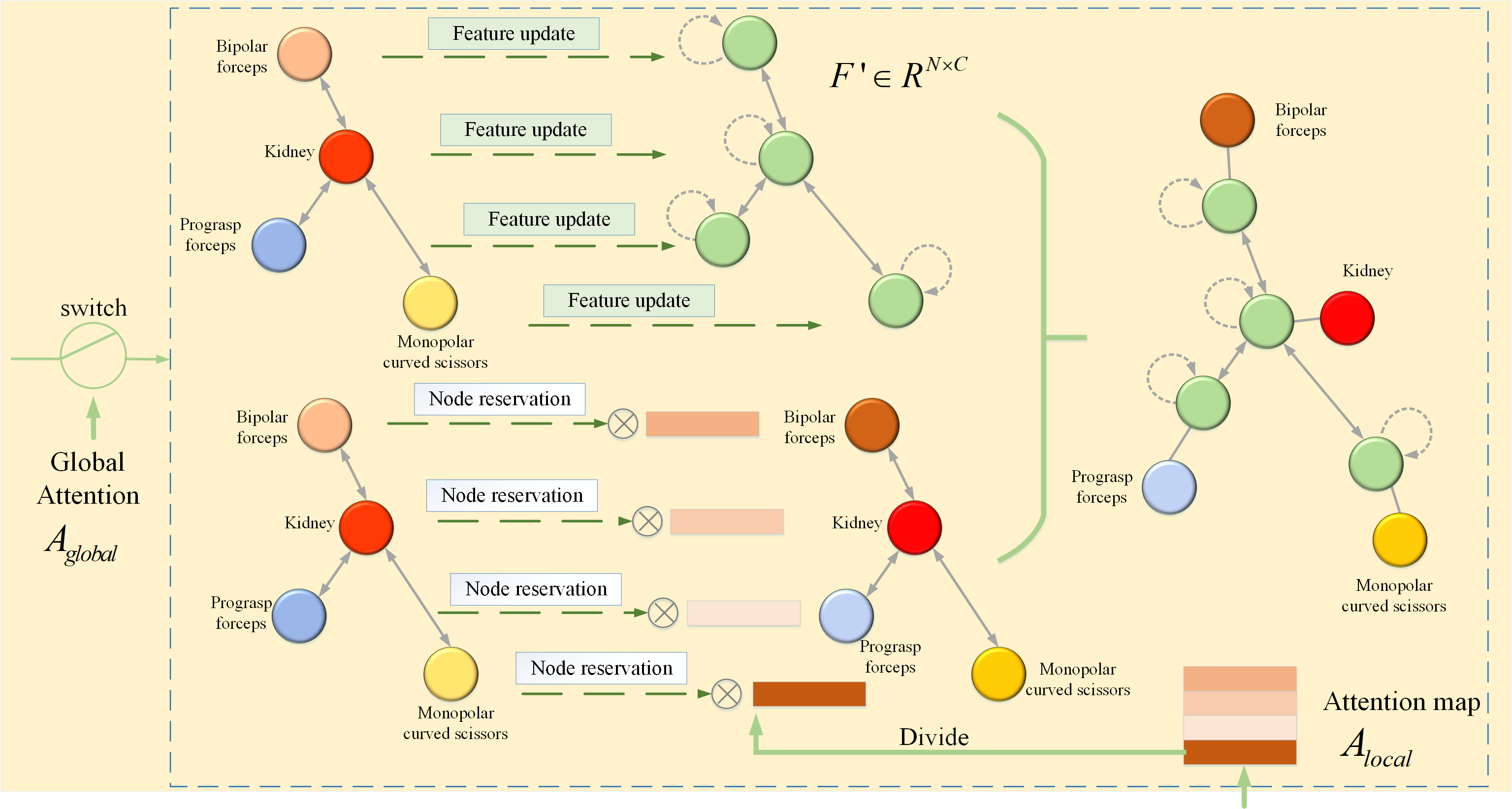}
    \caption{Detailed structure of the proposed RE module. The RE module will generate new node embeddings by updating features, while reserving the original node embeddings. Note that the embeddings of the graph nodes come from the features extracted by ResNet18 in the object detection areas. Global attention works in front of the RE module as a switch to decide whether to perform relational exploration, while local attention works inside the RE module to enhance the interactive node features.}
    \label{fig:RE_module}
\end{figure*}

\section{Methods}

\subsection{Overview}

Fig.~\ref{fig:pipeline} shows the schematic illustration of our surgical report generation network.  
Unlike existing surgical report generation methods taking a single image as the input, our method takes a surgical video as the input and then generates the surgical report for each video frame by exploring the interaction relations between tissues and surgical instruments. 
Specifically, given a video frame $I_t$, we take two adjacent video frames $I_{t-1}$ and $I_{t-2}$, and employ YOLOv5 \cite{yolo} as the object detector to detect objects from all three input video frames (i.e., $I_t$, $I_{t-1}$ and $I_{t-2}$). 
Then,  we devise a node tracking mechanism in our interaction perception (IP) module to further identify and append some missing nodes of the current video frame $I_t$ by leveraging the object detection results of the adjacent video frames.
Moreover, we apply a feature extractor (i.e., ResNet18 \cite{19} following previous work \cite{9}) to extract features of each identified node and devise a RE module to leverage the graph learning for learning the interactive relation between tissues and surgical instruments. 
More importantly, we devise an IP module to predict a global attention map to classify the interactive relation on the whole scene and predict a local attention map to identify the interactive relation on nodes to assist the graph learning at our RE module. 
After that, we concatenate features of all graph nodes of the RE module and pass the concatenation result into the M2 transformer~\cite{13} for predicting the output surgical report of the current video frame $I_t$. 

\subsection{Relational Exploration Module}

Recently,  due to the capability of modeling
non-Euclidean relationships among entities, GNNs have achieved promising performances on diverse applications including image classification~\cite{wang2018zero}, neural machine translation~\cite{marcheggiani2018exploiting}, social relationship understanding~\cite{wang2018deep}, and gesture recognition in robotic surgery~\cite{long2021relational}.
Motivated by this, we propose to model the interaction relation between tissues and surgical instruments via graph learning.
Fig.~\ref{fig:RE_module} shows the schematic illustration of the proposed RE module.

The node embeddings of our graph come from the feature maps $F $ extracted by ResNet18. 
RE module will update the embeddings as \begin{equation}
    F'= \sigma (\widetilde{D}^{-1/2} \widetilde{A} \widetilde{D}^{-1/2} F W),
\end{equation} 
where $ \widetilde{A} $ is the adjacency matrix of the undirected graph $\mathcal{G}$ with added self-connections, $\widetilde{D_{ii}} = \sum_{j}\widetilde{A}_{ij}$ , $W$ is a layer-specific trainable weight matrix and $\sigma ( )$ denotes an activation function (i.e. ReLU). 
By doing so, the representations of the interactions $F'$ between different nodes can be obtained, which can effectively improve the accuracy of the generated report. 

Preserving the inherent characteristic of object own is also of vital importance for this task. Because the node will exchange information with its connected nodes, the updated node embeddings are more inclined to interactive representation, which may dilute its object information. Especially for some core components, e.g., the node of tissues, it generally interacts with multiple objects, whose feature shall be disturbed by those multiple nodes. In this regard, we devise the node reservation operation, to simultaneously consider and model both inherent object representations and interaction information in the scene, which will facilitate subsequent text generation.

\subsection{Interaction Perception module}

Since the surgical instruments can be idle during a surgical video,  it is possible that the input surgical video has one or more video frames without any interaction relation between tissues and surgical tools.
In this regard, the surgical report generation performance degrades if there is no interaction at the current video frame and we still utilize our RE module to model the node relation.
To alleviate this issue, we develop an IP module to explicitly classify whether the current video frame has an interaction relation between tissues and surgical instruments. 


\textbf{Node Tracking.} To do so, our IP module first generates a complete scene graph for each video frame. However, since the graph of different frames of the surgical video may vary greatly, some key nodes may be missing in the scene graphs of some frames. 
To alleviate this issue, we devise a node tracking mechanism to utilize temporal information to continuously track key nodes among input adjacent video frames.
As shown in Fig. 2, our IP module utilizes the object detection results of each video frame to construct a scene graph $\mathcal{G}=\{\mathcal{V},\mathcal{E},\mathcal{R}\}$ with nodes $v_i \in \mathcal{V}$, edges $(v_i, r, v_j) \in \mathcal{E}$ and a relation $r \in \mathcal{R}$. Regarding robotic surgery, we believe that the surgical instruments that appear in different frames are constantly changing, but the surgical target needs to be continuously tracked. 
As shown in Fig. 2, we track the kidney node and the tracking length is set as three video frames. When a video frame has missing nodes, it cannot form a complete scene map. Our node tracking mechanism will add the missing nodes according to scene maps of previous adjacent frames. By doing so, we can obtain a complete scene map for the video frame to assist the subsequent surgical report generation. 

\textbf{Global attention maps.} 
The interactive relation reasoning is not required along the whole surgical sequence. The timesteps when the instruments are separate from the tissues (e.g., Preparation phase in the surgery), performing the interactive modeling via graph instead inevitably brings some interferences. In this regard, we propose to only invoke the RE module after observing the actual interaction in the whole scene globally. 
Specifically, once we obtain a complete scene graph of the current video frame $I_t$, we then obtain a feature map of each node of the scene graph by extracting deep features from the detected object corresponding to this scene graph node. 
Then, we concatenate features of all nodes of the scene graph of $I_t$, and then pass the concatenated features $F_{con}$ into a multi-layer perception block (see Fig. 2) to classify whether there is an interaction relation between tissues and surgical instruments in the whole scene of the video frame $I_t$.
Specifically, the multi-layer perception block applies three fully-connected layers on the concatenated features to obtain a global attention map $\mathcal{A}_{global}$, which has only a scalar, and the scalar value can be 0 or 1.
\begin{equation}
    \label{Eq-global-attention}
    \mathcal{A}_{global} = \Phi_1(\Phi_2(\Phi_3(F_{con}))) \ ,
\end{equation},
where $\Phi_1$, $\Phi_2$, and $\Phi_3$ denote three fully-connected layers. 
Apparently, our $\mathcal{A}_{global}$ represents whether there is an interaction result between tissues and surgical instruments in the whole scene.

\textbf{Local attention map.} 
Although there are interaction relations in a whole scene view, we find that not all graph nodes are involved in these interaction relations, and the surgical report tends to focus on these involved nodes and ignore these idle nodes, which are not involved in interactions. 
In this regard, apart from predicting a global attention map for the whole scene, our IP module also predicts a local attention map to assign different weights to different graph nodes, thereby boosting the surgical report generation. 

Specifically, we apply another three fully-connected layers on $F_{con}$ to generate a local attention map $\mathcal{A}_{local}$, which is a vector with $N$ ($N$ represents the number of nodes at the graph of the RE module) elements. 
\begin{equation}
    \label{Eq-global-attention}
    \mathcal{A}_{local} = \Phi_4(\Phi_5(\Phi_6(F_{con}))) \ ,
\end{equation},
where $\Phi_4$, $\Phi_5$, and $\Phi_6$ denote three fully-connected layers. 
Apparently, our $\mathcal{A}_{local}$ represents whether there is an interaction result for all graph nodes.

\subsection{Implementation Details}

All experiments were implemented on PyTorch and trained on an NVIDIA GeForce RTX 2080 Ti GPU with 11 GB memory. For object detection, BCEWithLogits loss and CIoU loss are empirically applied to compute the loss function. 
As for the detected ROI areas, all image patches are resized to 224×224 before passing them into ResNet18. 
For the training caption generation part, we adopt the CE loss and Adam optimizer \cite{24} with a learning rate of 0.00006. 
The learning rate is then decayed by an exponential function with a factor of 0.8 for every 10 epochs. 
All models were trained with 80 epochs. The batch size is set to 50. Following previous works \cite{9}, all words in each surgical report will be changed to be lowercase, and punctuation is also removed.

\begin{table*}[t]
\caption{Quantitative comparisons of our network and state-of-the-art surgical report generation method on  ON THE MICCAI ROBOTIC CHALLENGE DATASET. “-”  denotes that the results are not available.}
\label{t2}
\vspace{-3mm}
\begin{center}
\begin{tabular}{c|c|c|c|c|c|c|c}
\hline
Models & BLEU-1$\uparrow$ & BLEU-2$\uparrow$ & BLEU-3$\uparrow$ & BLEU-4$\uparrow$ & METEOR$\uparrow$ & ROUGE$\uparrow$ & CIDEr$\uparrow$\\
\hline
Xu et al \cite{8} & 0.5228 & 0.4730 & 0.4262 & 0.3861 & \textbf{0.4567} & 0.6495 & 2.2598\\
V-SwinMLP-TranCAP \cite{31} & - & - & - & 0.4230 & 0.3780 & - & 2.6630\\
CIDA \cite{9} & 0.6246 & 0.5624 & 0.5117 & 0.4720 & 0.38 & 0.6294 & 2.8548\\
\hline
Our method & \textbf{0.6994} & \textbf{0.6352} & \textbf{0.5807} & \textbf{0.5332} & 0.41 & \textbf{0.7038} & \textbf{3.9006}\\
\hline
\end{tabular}
\end{center}\end{table*}

\section{Experiments and Results}

\subsection{Dataset}

We evaluate the effectiveness of our method on a widely-used benchmark dataset~\cite{25} from 2018 MICCAI Robotic Instrument Segmentation Endoscopy Vision Challenge. 
This dataset contains 15 robotic nephrectomy procedures captured on the da Vinci X or Xi system and each video (15 videos in total) has 149 frames with a spatial resolution of 1280×1024. 
The surgical reports contained a total of 11 interactive relationships, including manipulating, grasping, retracting, cutting, cauterizing, looping, suctioning, clipping, ultrasound sensing, stapling, and suturing. Besides, 9 objects appeared in the dataset, and they are the kidney and 8 instruments (monopolar curved scissors, bipolar forceps, prograsp forceps, clip applier, suction, ultrasound probe, stapler, and large needle driver). These interactive relationships and object information together form scene graph representations, which are important elements of natural language description. 
Following the previous works \cite{9}, we remove the 13th sequence due to its few interactions, and utilize 14 surgical videos for training and validation. 
And the 1st, 5th, 6th surgical videos are utilized for validation, and the remaining 11 videos are for training different methods to conduct a fair comparison.

\subsection{Evaluation Metrics}
To quantitatively verify the effectiveness of the proposed methods, seven commonly-used metrics for image captioning are introduced. They are BLEU-1 \cite{26}, BLEU-2 \cite{26}, BLEU-3 \cite{26}, BLEU-4 \cite{26}, METEOR \cite{27}, ROUGE \cite{28}, and CIDEr \cite{29}. 
In general, a better surgical report generation method should have larger scores of all seven metrics. 


\subsection{Comparisons Against State-of-the-art Methods}

\begin{figure}
    \centering
    \includegraphics[width=0.5\textwidth]{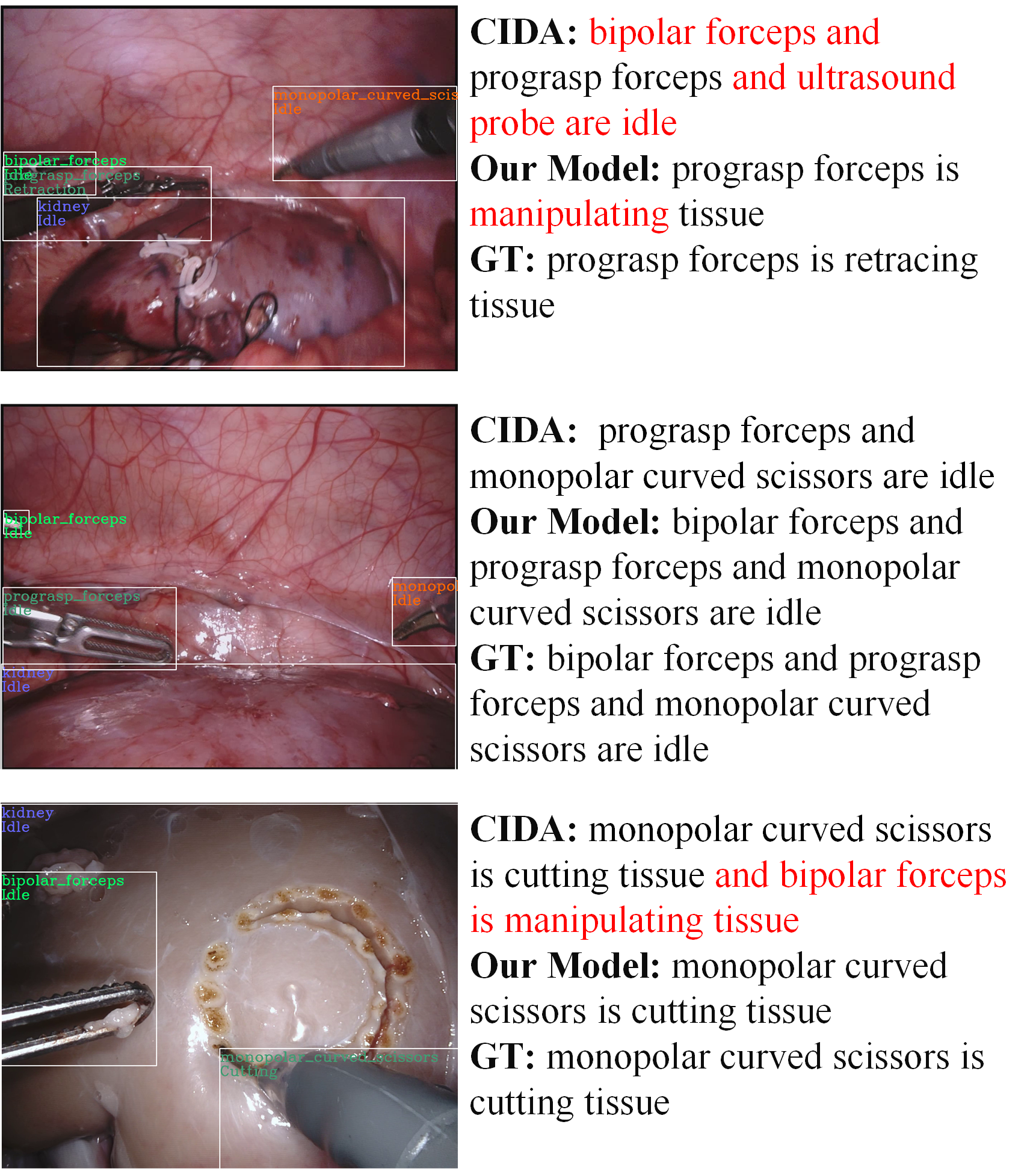}
    \caption{Visual comparisons of the surgical report generation results produced by our network and the most recent CIDA. Red words indicate that the prediction was not correct.}
    \label{fig:SOTA}
\end{figure}

\begin{table*}[t]
\caption{Ablation study on our RE and IP modules of our network. }
\label{table:Ablation_study_RE_IP}
\vspace{-1mm}
\begin{center}
\begin{tabular}{c|c|c|c|c|c|c|c|c|c}
\hline
\multicolumn{3}{c|}{} & \multicolumn{7}{c}{Evaluation metrics}\\
\hline
 Method& RE & IP & BLEU-1$\uparrow$& BLEU-2$\uparrow$& BLEU-3$\uparrow$& BLEU-4$\uparrow$& METEOR$\uparrow$& ROUGE$\uparrow$& CIDEr$\uparrow$\\
\hline
Basic & \usym{2715} & \usym{2715} & 0.6308 & 0.5686 & 0.5147 & 0.4673 & 0.36 & 0.6135 & 2.8600\\
\hline
Basic+RE & \usym{2713} & \usym{2715} & 0.6667 & 0.5961 & 0.5402 & 0.4941 & 0.38 & 0.6583 & 3.4842\\
\hline
Basic+IP & \usym{2715} & \usym{2713} & 0.6781 & 0.6118 & 0.5662 & 0.5288 & 0.38 & 0.6662 & 3.6988\\
\hline
Basic+RE+IP (Our method) & \usym{2713} & \usym{2713} & \textbf{0.6994} & \textbf{0.6352} & \textbf{0.5807} & \textbf{0.5332} & \textbf{0.41} & \textbf{0.7038} & \textbf{3.9006}\\

\hline
\end{tabular}
\end{center}
\end{table*}

\begin{table*}[!htbp]
\caption{Ablation study on the node tracking mechanism, the global attention map, and the local attention map in our IP module. NT, GA, and LA denote the node tracking mechanism, the global attention map, and the local attention map, respectively.}
\label{table:Ablation_study_IP}
\vspace{-1mm}
\begin{center}
\begin{tabular}{c|c|c|c|c|c|c|c|c|c|c|c}
\hline
\multicolumn{5}{c|}{} & \multicolumn{7}{c}{Evaluation metrics}\\
\hline
\multirow{2}{*}{Method} & \multirow{2}{*}{Basic+RE} &  \multicolumn{3}{c|}{IP module} &  \multirow{2}{*}{BLEU-1$\uparrow$} & \multirow{2}{*}{BLEU-2$\uparrow$} & \multirow{2}{*}{BLEU-3$\uparrow$} &  \multirow{2}{*}{BLEU-4$\uparrow$} & \multirow{2}{*}{METEOR$\uparrow$} & \multirow{2}{*}{ROUGE$\uparrow$} & \multirow{2}{*}{CIDEr$\uparrow$} \\ 
\cline{3-5} & &NT & GA& LA & & & & & & &\\
\hline
M1&\usym{2713} & \usym{2715} & \usym{2715} & \usym{2715} & 0.6667 & 0.5961 & 0.5402 & 0.4941 & 0.38 & 0.6583 & 3.4842\\
\hline
M2&\usym{2713} & \usym{2713} & \usym{2715} & \usym{2715} & 0.6748 & 0.6181 & 0.5728 & 0.5367 & 0.39 & 0.6616 & 3.6071\\
\hline
M3&\usym{2713} & \usym{2715} & \usym{2713} & \usym{2715} & 0.6773 & 0.6162 & 0.5764 & 0.5373 & 0.39 & 0.6712 & 3.6753\\
\hline
M4&\usym{2713} & \usym{2715} & \usym{2715} & \usym{2713} & 0.6786 & 0.6173 & 0.5675 & 0.5245 & 0.39 & 0.6683 & 3.8567\\
\hline
M5&\usym{2713} & \usym{2713} & \usym{2713} & \usym{2715} & 0.6818 & 0.6263 & 0.5711 & \textbf{0.5340} & 0.40 & 0.6705 & 3.7743\\
\hline
Our method&\usym{2713} & \usym{2713} & \usym{2713} & \usym{2713} & \textbf{0.6994} & \textbf{0.6352} & \textbf{0.5807} & 0.5332 & \textbf{0.41} & \textbf{0.7038} & \textbf{3.9006}\\

\hline
\end{tabular}
\end{center}
\end{table*}

\textbf{Quantitative comparisons.} \ We compare our method against state-of-the-art surgical report generation methods based on deep learning, which are Xu et al. \cite{8}, V-SwinMLP-TranCAP \cite{31}, and CIDA \cite{31}. 
Among three compared methods, we can find that CIDA has the best performance of BLEU-1, BLEU-2, BLEU-3, BLEU-4, and CIDEr. They are 0.6246, 0.5624, 0.5117, 0.4720, and 2.8548, while Xu et al. has the best performance of METEOR (0.4567) and ROUGE (0.6495).
Compared to the best performing existing methods, our network obtains a BLEU-1 improvement of 11.97 \%, a BLEU-2 improvement of 12.94\%, a BLEU-3 improvement of 13.48\%, a BLEU-3 improvement of 12.96\%, a ROUGE improvement of 8.36\%, and a CIDEr improvement of 36.63\%, respectively. 
Specifically, our method has largest BLEU-1, BLEU-2, BLEU-3, BLEU-4, ROUGE, and CIDEr scores, and they are 0.6994, 0.6352, 0.5807, 0.5332, 0.7038, 3.9006.  
Moreover, our method takes the second rank of METEOR score, and our METEOR score is 0.4100, which is slightly smaller than the best one (0.4567).
It indicates that our network can generate more accurate surgical reports than compared state-of-the-art methods.

Note that the metric CIDEr inherently captures the sentence similarity using the notions of grammaticality, saliency, importance and accuracy (precision and recall). 
Hence the CIDEr score is highly consistent with the consensus of human assessments. 
From the Table I, we can find that our method has gained a huge improvement on CIDEr (36.63\%).
It indicates that the report generated by our method is closer to the annotated report provided by the doctors than that of other state-of-the-art models.

\textbf{Visual comparisons.} \
Fig.~\ref{fig:SOTA} visually compares the generated surgical report of our method and CIDA \cite{9}.
Apparently, our method can more accurately predict the interaction operations of the surgical report since our method explicitly learns the interactive relation via graph learning.
Taking the first image of Fig.~\ref{fig:SOTA} as an example, CIDA tends to predict that the ultrasound probe is idle, and our method can correctly predict the interactive relation between the prograsp forceps and tissues.
Regarding the second image in Fig.~\ref{fig:SOTA}, we can find that CIDA missed that the bipolar forceps are also idle as the prograsp forceps and monopolar curved scissors. This is because the attention maps of the IP module enable our method can identify all instruments of the input surgical videos. 
Regarding the 3rd image, CIDA wrongly predicts an interactive relation between bipolar forceps and kidney. By exploring the relationship between different nodes, the interactive instrument is correctly estimated in the generated report of our method.


\subsection{Ablation Analysis}

\textbf{Effectiveness of our RE module and IP module.} \ We further conduct ablation study experiments to validate the effectiveness of our RE module and our IP module. 
To do so, we construct a baseline (denoted as ``Basic'') by removing our RE module and our IP module from our network, and then add the RE module and the IP module into ``Basic'' to build another two networks, which are denoted as ``Basic+RE'' and ``Basic+IP''.
As shown in Table~\ref{table:Ablation_study_RE_IP}, ``Basic+RE'' and ``Basic+IP'' has a better metric performance than ``Basic'' in terms of all seven metrics, which demonstrates that the RE module and the IP module can improve the surgical report generation performance of our method. 
Moreover, by adding the RE module and the IP module together, our method can generate a more accurate surgical report due to our superior metric results over ``Basic+RE'' and ``Basic+IP''.

\textbf{Effectiveness of key components in IP module.} \ As shown in Fig.~\ref{fig:pipeline}, we in our IP module devise has a node tracking mechanism to adding the possible missing nodes of the scene graph of the input video frame, and a global attention map on the whole scene, and a local attention map on the graph nodes of the RE module.
To further evalute the effectiveness of the node tracking mechanism, and the global attention map, and the local attention map, we conduct another ablation study experiment.
Here we construct five baseline networks (see M1 to M5 of Table~\ref{table:Ablation_study_IP}), which are reconstructed by only modifying the IP module of our network. It means that all these five baseline networks are build on   ``Basic+RE'' (see Table~\ref{table:Ablation_study_RE_IP}).

Table~\ref{table:Ablation_study_IP} reports the BLEU-1, BLEU-2, BLEU-3, BLEU-4, METEOR, ROUGE, and CIDEr scores of our method and five baseline networks.
Apparently, M2, M3, and M4 has larger scores on all seven metrics than M1, which means that each component of the node tracking mechanism, the global attention, and the local attention in our IP module enables our network to generate a more accurate surgical report.
Moreover, exploring both the node tracking mechanism and the global attention map together (i.e., M5) in our IP module incurs a performance gain than that with only the node tracking mechanism (i.e., M2), due to the superior performance of M5 over M2. It demonstrates that the global attention on the whole scene of the input video frame on the RE module helps our method to generate a more accurate surgical report. 
More importantly, our method has larger BLEU-1, BLEU-2, BLEU-3, BLEU-4, METEOR, ROUGE, and CIDEr scores than M5, which indicates that incorporating the local attention map on graph nodes in our IP module also boost the surgical report generation performance of our network.

\section{Conclusion and Future Work}

This work presents a new surgical report generation method by exploring the interactive relation between tissues and instruments via graph learning.
Our key idea is to devise a RE module to leverage temporal information to model interactive relations, and devise an IP module to assist the graph learning in RE module.
The IP module has a node tracking system can identify and append missing nodes of the current video frame for assisting the graph network construction in RE module.
Moreover, the IP module generates a global attention map to indicate the existence of the interactive relation on the whole scene of the current video frame, and a local attention map to perceive the interactive relation on each graph node of the RE module.
By doing so, the graph updating in the RE module will be more accurate, thereby enhancing the surgical report generation accuracy.
Experimental results on 2018 MICCAI Endoscopic Vision Challenge Dataset show that our network clearly outperform existing state-of-the-art surgical report generation methods.
In the future, we plan to consider incorporating multi-modality information, such as kinematics to facilitate report generation. In addition, we also plan to collect more data to extend our method to multiple interaction points of multiple human organs.

\begin{refcontext}[sorting = none]
\printbibliography
\end{refcontext}

\end{document}